\documentclass{article}

\usepackage{microtype}
\usepackage{graphicx}
\usepackage{subfigure}
\usepackage{booktabs} 

\usepackage{lmodern}

\usepackage{hyperref}

\usepackage{amsmath,amsthm,amssymb}
\usepackage{thmtools}
\usepackage{xcolor}
\usepackage{thm-restate}
\usepackage{mathtools}
\usepackage{tikz}
\usetikzlibrary{shapes,decorations,arrows,calc,arrows.meta,fit,positioning}
\tikzset{
    -Latex,auto,node distance =0.1 cm and 1 cm,semithick,
    state/.style ={circle, draw, minimum width = 0.7 cm},
    point/.style = {circle, draw, inner sep=0.04cm,fill,node contents={}},
    bidirected/.style={Latex-Latex,dashed},
    el/.style = {inner sep=2pt, align=left, sloped}
}

\RequirePackage{algorithm}
\RequirePackage{algorithmic}

\declaretheorem{theorem}

\declaretheorem{lemma}

\declaretheorem{remark}

\declaretheorem[style=definition]{objective}
\newenvironment{proofidea}{
    \proof}{\endproof}

\newcommand{\R}{\mathbb{R}}

\newcommand{\E}[2][\space]{\mathbb{E}_{#1} \left[ #2 \right] }

\newcommand{\defeq}{\vcentcolon=}

\DeclareMathOperator*{\argmax}{arg\,max}
\DeclareMathOperator*{\argmin}{arg\,min}

\DeclareMathOperator{\AO}{AO}

\usepackage[accepted]{icml2020_style_final/icml2020}

\icmltitlerunning{Causal Strategic Linear Regression}

\begin{document}

\twocolumn[
\icmltitle{Causal Strategic Linear Regression}




\begin{icmlauthorlist}
\icmlauthor{Yonadav Shavit}{har}
\icmlauthor{Benjamin L. Edelman}{har}
\icmlauthor{Brian Axelrod}{sta}
\end{icmlauthorlist}

\icmlaffiliation{har}{Harvard School of Engineering and Applied Sciences,
Cambridge, MA, USA}
\icmlaffiliation{sta}{Stanford Computer Science Department, Palo Alto, CA, USA}

\icmlcorrespondingauthor{Yonadav Shavit}{yonadav@g.harvard.edu}

\icmlkeywords{Strategic classification, Mechanism design}

\vskip 0.3in
]



\printAffiliationsAndNotice{}  

\begin{abstract}
	In many predictive decision-making scenarios, such as credit scoring and academic testing, a decision-maker must construct a model that accounts for agents' propensity to ``game'' the decision rule by changing their features so as to receive better decisions. Whereas the strategic classification literature has previously assumed that agents' outcomes are not causally affected by their features (and thus that strategic agents' goal is deceiving the decision-maker), we join concurrent work in modeling agents' outcomes as a function of their changeable attributes. As our main contribution, we provide efficient algorithms for learning decision rules that optimize three distinct decision-maker objectives in a realizable linear setting: accurately predicting agents' post-gaming outcomes (\emph{prediction risk minimization}), incentivizing agents to improve these outcomes (\emph{agent outcome maximization}), and estimating the coefficients of the true underlying model (\emph{parameter estimation}). 
Our algorithms circumvent a hardness result of \citet{miller2020} by allowing the decision maker to test a sequence of decision rules and observe agents' responses, in effect performing causal interventions through the decision rules.
\end{abstract}

\section{Introduction}

As individuals, we want algorithmic transparency in decisions that affect us.
Transparency lets us audit models for fairness and correctness, and allows us to
understand what changes we can make to receive a different decision. Why,
then, are some models kept hidden from the view of those subject to their
decisions?

Beyond setting-specific concerns like intellectual property theft or
training-data extraction, the canonical answer is that transparency would allow
strategic individual \emph{agents} to ``game'' the model. These individual
agents will act to change their features to receive better
\emph{decisions}. An accuracy-minded decision-maker, meanwhile, chooses a decision rule based on how well it predicts individuals' true future \emph{outcomes}. Strategic agents, the
conventional wisdom goes, make superficial changes to their features that will lead to them receiving more desirable decisions, but these feature changes will
not affect their true outcomes, reducing the decision rule's accuracy and
harming the decision-maker.
The field of strategic classification \cite{hardt2016} has until
recently sought to design algorithms that are robust to such superficial
changes. At their core, these algorithms treat transparency as a reluctant concession
and propose ways for decision-makers to get by nonetheless.

But what if decision-makers could \emph{benefit} from transparency? What if in
some settings, gaming could help accomplish the decision-makers' goals, by
causing agents to truly improve their outcomes without loss of predictive
accuracy?

Consider the case of car insurance companies, who wish to choose a pricing
decision rule that charges a customer in line with the expected costs that customer will incur through accidents.
Insurers will often charge lower prices to drivers who have completed a
``driver's ed'' course, which teaches comprehensive driving skills. In response, new drivers will often complete such courses to reduce their insurance
costs. One view may be that only \emph{ex ante} responsible drivers seek out 
such courses, and that were an unsafe driver to complete such a course it would not affect their expected cost of car accidents.

But another interpretation is that drivers in these courses learn
safer driving practices, and truly become safer drivers \emph{because} they take
the course. In this case, a car insurer's decision rule \emph{remains
predictive} of accident probability when agents strategically game the decision rule, while also incentivizing the population of
drivers to act in a way that truly causes them to have fewer accidents, which means the insurer needs to pay out fewer reimbursements.

This same dynamic appears in many decision settings where the
decision-maker has a meaningful stake in the true future outcomes of its subject
population, including credit scoring, academic testing, hiring, and online
recommendation systems. In such scenarios, given the right decision rule,
decision-makers can \emph{gain} from transparency.

\begin{figure}\label{fig:causal_graph}\centering
\begin{tikzpicture}
\node[state] (w) [dashed] at (0,0) {$\omega$};
    \node[state] (x) [right =of w] {$x$};
    \path (w) edge (x);
    \node[state] (y) [below right =of x] {$y$};
    \path (x) edge (y);
    \node[state] (wstar) [below left=of y] {$\omega^*$};
	\path (wstar) edge (y);\end{tikzpicture}
\caption{A causal graph illustrating that by intervening on the decision rule $\omega$, a decision-maker can incentivize a change in $x$, enabling them to learn about how the agent outcome $y$ is caused. We omit details of our setting for simplicity.}
\end{figure}
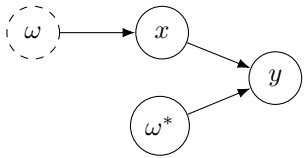

But how can we find such a decision rule that maximizes agents' outcomes if we do not know the effects of agents' feature-changing actions?
In recent work, \citet{miller2020} argue that finding such ``agent
outcome''-maximizing decision rules requires solving a non-trivial causal inference problem.
As we illustrate in Figure \ref{fig:causal_graph}, the decision rule affects the agents' features, which causally
affect agents' outcomes, and recovering these relationships from observational
data is hard. We will refer to this setting as ``causal strategic
learning'', in view of the causal impact of the decision rule on agents' outcomes.

The core insight of our work is that while we may not know how agents will
respond to any arbitrary decision rule, they will naturally respond to any particular rule we pick.
Thus, as we test different decision rules and observe strategic agents' responses
and true outcomes, we can improve our decision rule over time. In the language of
causality, by choosing a decision rule we are effectively launching an intervention
that allows us to infer properties of the causal graph, circumventing
the hardness result of \citet{miller2020}.

In this work, we introduce the setting of causal strategic linear regression in the realizable case and with norm-squared agent action costs.
We propose algorithms for efficiently optimizing three possible objectives that a decision-maker may care about. \emph{Agent
outcome maximization} requires choosing a decision rule that will result in the highest expected outcome of an agent who games that rule. \emph{Prediction risk
minimization} requires choosing a decision rule that accurately predicts agents' outcomes, even under agents' gaming in response to that same decision rule. \emph{Parameter estimation} involves accurately estimating the parameters of the true causal outcome-generating linear model.
We show that these may be mutually non-satisfiable, and our algorithms maximize each objective independently (and jointly when possible).

Additionally, we show that omitting unobserved yet outcome-affecting features from the decision rule has major consequences for causal strategic learning.
Omitted variable bias in classic linear regression leads a learned predictor to reward
non-causal visible features which are correlated with hidden
causal features \cite{greene2003econometric}. In the strategic
case, this backfires, as an agent's action may change a visible feature without changing the hidden feature, thereby breaking this correlation, and undermining a naively-trained
predictor. All of our methods are designed to succeed even when actions
break the relationships between visible proxies and hidden causal features. Another common characteristic of real-life strategic learning scenarios is that in order to game one feature, an agent may need to take an action that also perturbs other features. We follow \citet{kleinberg2019} in modeling this by having agents take actions in \emph{action space} which are mapped to features in \emph{feature space} by an \emph{effort conversion matrix}.

As much of the prior literature has focused on the case of binary
classification, it's worth meditating on why we focus on regression. Many
decisions, such as loan terms or insurance premiums, are not binary
``accept/reject''s but rather lie somewhere on a continuum based on a prediction
of a real-valued outcome. Furthermore, many ranking decisions, like which
ten items to recommend in response to a search query, can
instead be viewed as real-valued predictions that are post-processed into an
ordering.

\subsection{Summary of Results}
In Section \ref{sec:setting}, we introduce a setting
for studying the performance of linear models that make real-valued decisions about strategic agents. 
Our methodology incorporates the realities that agents' actions may causally affect their eventual
outcomes, that a decision-maker can only observe a subset of agents' features,
and that agents' actions are constrained to a subspace of the feature space. We
assume no prior knowledge of the agent feature distribution or of the actions
available to strategic agents, and require no knowledge of the true outcome
function beyond that it is itself a noisy linear function of the features.

In Section \ref{sec:outcome}, we propose an algorithm for efficiently learning a decision rule that will maximize agent outcomes. The algorithm involves publishing a decision rule corresponding to each basis vector; it is non-adaptive, so each decision rule implemented in parallel on a distinct portion of the population.

In Section \ref{sec:risk}, we observe that under certain checkable conditions the prediction risk objective can be minimized using gradient-free convex optimization techniques. We also provide a useful decomposition of prediction risk, and suggest how prediction risk and agent outcomes may be jointly optimized.

In Section \ref{sec:parameter_estimation}, we show that in the case where all features that causally affect the outcome are visible to the decision-maker, one can substantially improve the estimate of the true model parameters governing the outcome. At a high level, this is because by incentivizing agents to change their features in certain directions, we can improve the conditioning of the second moment matrix of the resulting feature distribution.


\subsection{Related Work}

This paper is closely related to several recent and concurrent papers that study different aspects of (in our parlance) causal strategic learning. Most of these works focus on one of our three objectives.

\paragraph{Agent outcomes.} Our setting is partially inspired by \citet{kleinberg2019}. In their setting, as in ours, an agent chooses an action vector in order to maximize the score they receive from a decision-maker. The action vector is mapped to a feature vector by an \emph{effort conversion matrix}, and the decision-maker publishes a mechanism that maps the feature vector to a score. However, their decision-maker does not face a learning problem: the effort conversion matrix is given as input, agents do not have differing initial feature vectors, and there is no outcome variable. Moreover, there are no hidden features. In a variation on the agent outcomes objective, their decision-maker's goal is to incentivize agents to take a particular action vector. Their main result is that whenever a monotone mechanism can incentivize a given action vector, a linear mechanism suffices. \citet{alon2020} analyze a multi-agent extension of this model.

In another closely related work, \citet{miller2020} bring a causal perspective \cite{pearl2000,peters2017} to the strategic classification literature. Whereas prior strategic classification works mostly assumed agents' actions have no effect on the outcome variable and are thus pure \emph{gaming}, this paper points out that in many real-life strategic classification situations, the outcome variable is a descendant of some features in the causal graph, and thus actions may lead to genuine \emph{improvement} in agent outcomes. Their main result is a reduction from the problem of orienting the edges of a causal graph to the problem of finding a decision rule that incentivizes net improvement. Since orienting a causal graph is a notoriously difficult causal inference problem given only observational data, they argue that this provides evidence that incentivizing improvement is hard. In this paper we we point out that improving agent outcomes may not be so difficult after all because the decision-maker does not need to rely only on observational data---they can perform causal interventions through the decision rule.

\citet{haghtalab2020} study the agent outcomes objective in a linear setting that is similar to ours. A significant difference is that, while agents do have hidden features, they are never incentivized to change their hidden features because there is no effort conversion matrix. This, combined with the use of a Euclidean norm action cost (we, in contrast, use a Euclidean squared norm cost function), makes finding the optimal linear regression parameters trivial. Hence, they mainly focus on approximation algorithms for finding an optimal linear \emph{classifier}.

\citet{tabibian2020} consider a variant of the agent outcomes objective in a classification setting: the outcome is only ``realized'' if the agent receives a positive classification, and the decision-maker pays a cost for each positive classification it metes out. The decision-maker knows the dependence of the outcome variable on agent features a priori, so there is no learning. 

\paragraph{Prediction risk.}  \citet{perdomo2020} define \emph{performative prediction} as any supervised learning scenario in which the model's predictions cause a change in the distribution of the target variable. This includes causal strategic learning as a special case. They analyze the dynamics of \emph{repeated retraining}---repeatedly gathering data and performing empirical risk minimization---on the prediction risk. They prove that under certain smoothness and strong convexity assumptions, repeated retraining (or repeated gradient descent) converges at a linear rate to a near-optimal model. 

\citet{liu2020} introduce a setting where each agent responds to a classifier by intervening directly on the outcome variable, which then affects the feature vector in a manner depending on the agent's population subgroup membership. 

\paragraph{Parameter estimation.} \citet{bechavod2020} study the effectiveness of repeated retraining at optimizing the parameter estimation objective in a linear setting. Like us, they argue that the decision-maker's control over the decision rule can be conducive to causal discovery. Specifically, they show that if the decision-maker repeatedly runs least squares regression (with a certain tie-breaking rule in the rank-deficient case) on batches of fresh data, the true parameters will eventually be recovered. Their setting is similar to ours but does not include an effort conversion matrix.

\begin{center}
\rule{8em}{1px}
\end{center}

\paragraph{Non-causal strategic classification.} 
Other works on strategic classification are \emph{non-causal}---the decision-maker's rule has no impact on the outcome of interest.
The primary goal of the decision-maker in much of the classic strategic classification literature is robustness to gaming; the target measure is typically prediction risk. Our use of a Euclidean squared norm cost function is shared by the first paper in a strategic classification setting \cite{bruckner2011}. Other works use a variety of different cost functions, such as the \emph{separable} cost functions of \citet{hardt2016}. The online setting was introduced by \citet{dong2018} and has also been studied by \citet{chen2020}, both with the goal of minimizing ``Stackelberg regret''.\footnote{See \citet{bambauer2018} for a discussion of online strategic classification from a legal perspective.} A few papers \cite{milli2019, hu2019} show that accuracy for the decision-maker can come at the expense of increased agent costs and inequities. \citet{braverman2020} argue that random classification rules can be better for the decision-maker than deterministic rules.

\paragraph{Economics.} Related problems have long been studied in information economics, specifically in the area of contract theory \cite{salanie2005, laffont2002}. 
In \emph{principal-agent problems} \cite{holmstrom1979, grossman1983, holmstrom1991,ederer2018}, also known as \emph{moral hazard} or \emph{hidden action} problems, a decision-maker (called the \emph{principal}) faces a challenge very similar to the agent outcomes objective. Notable differences include that the decision-maker can only observe the outcome variable, and the decision-maker must pay the agent. 
In a setting reminiscent of strategic classification, \citet{frankel2020} prove that the fixed points of retraining can be improved in terms of accuracy if the decision-maker can commit to underutilizing the available information. \citet{ball2020} introduces a three-party model in which an intermediary scores the agent and a decision-maker makes a decision based on the score.

\paragraph{Ethical dimensions of strategizing.} \citet{ustun2019} and \citet{venkatasubramanian2020philosophical} argue that it is normatively good for individuals subject to models to have “recourse”: the ability to induce the model to give a desired prediction by changing mutable features. \citet{ziewitz2019} discusses the shifting boundaries between morally ``good'' and ``bad'' strategizing in the context of search engine optimization.

\paragraph{Other strategic linear regression settings.}
A distinct literature on strategic variants of linear regression \cite{perote2004,dekel2010,chen2018, ioannidis2013, cummings2015} studies settings in which agents can misreport their $y$ values to maximize their privacy or the model's prediction on their data point(s).

\section{Problem Setting}
\label{sec:setting}
\begin{figure}[ht]
    \centering
    \includegraphics[width=\linewidth, trim= 60 40 40 0, clip]{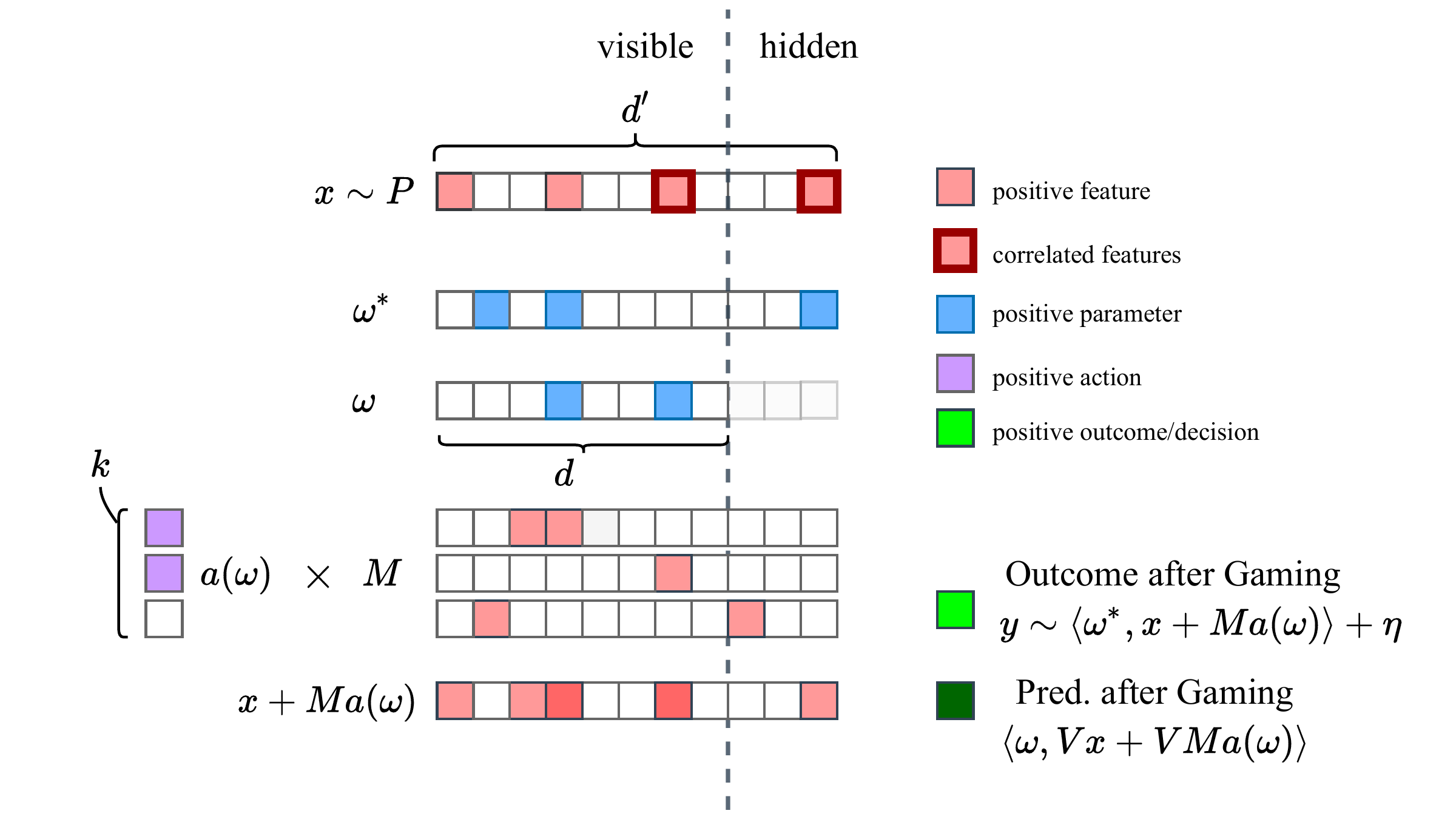}
    \caption{
        Visualization of the linear setting. Each box corresponds to a
        real-valued scalar, with value indicated by shading. The two boxes with dark red outlines represent features that are correlated in the initial feature distribution $P$.
        }
    \label{fig:setting}
\end{figure}

Our setting is defined by the interplay between two parties: \emph{agents},
who receive decisions based on their features, and a
\emph{decision-maker}, who chooses the decision rule that determines these decisions.\footnote{In
the strategic classification literature, these are occasionally referred to as
the ``Jury'' and ``Contestant''.}
We visualize our setting in Figure \ref{fig:setting}.

Each agent is described by a feature vector $x \in \R^{d'}$,\footnote{For simplicity of notation, we implicitly use homogeneous coordinates: one feature is 1 for all agents. The matrices $V$ and $M$, defined in this section, are such that this feature is visible to the decision-maker and unperturbable by agents.
}
initially drawn from a distribution $P \in \Delta(\R^{d'})$ over the
feature-space with second moment matrix $\Sigma = \E[x \sim P]{x x^T}$.
Agents can choose an action vector $a \in \R^k$ to change their features from
$x$ to $x_g$, according to
the following update rule: $x_g = x + Ma$ where the \emph{effort conversion
matrix} $M \in \R^{d' \times k}$ has an $(i, j)$th entry corresponding
to the change in the $i$th feature of $x$ as a result of spending one unit of effort along the $j$th direction of the action space. Each action dimension can affect multiple features
simultaneously. For example, in the context of car insurance, a prospective
customer's action might be ``buy a new car'', which can increase both the
safety rating of the vehicle and the potential financial loss from an
accident. The car-buying action might
correspond to a column $M_1 = (2, 10000)^T$, in which the two entries represent the
action's marginal impact on the car's safety rating and
cost-to-refund-if-damaged respectively. $M$ can be rank-deficient, meaning some feature
directions cannot be controlled independently through any action.

Let $y$ be a random variable representing an agent's true outcome, which we
assume is decomposable into a noisy linear combination of the features $y
\defeq \langle \omega^*, x_g \rangle + \eta$, where $\omega^* \in \R^{d'}$ is
the true \emph{parameter vector},
and $\eta$ is a subgaussian noise random variable with variance $\sigma$. 
Note that $\omega^*_i$ can be understood as the causal effect of a
change in feature $i$ on the outcome $y$, in expectation. Neither the
decision-maker nor the agent knows $\omega^*$.

To define the decision-maker's behavior, we must introduce an important aspect
of our setting: the decision-maker never observes an agent's complete feature
vector $x_g$, but only a subset of of those features $Vx_g$, where $V$ is a
diagonal projection matrix with $1$s for the $d$ visible features and $0$s for the hidden
features.

Now, our decision-maker assigns decisions $\langle \omega, Vx_g
\rangle $, where $\omega \in \R^{d'}$ is the \emph{decision rule}. Note that because the
hidden feature dimensions of $\omega$ are never used, we will define them to be
$0$, and thus $\omega$ is functionally defined in the $d$-dimensional visible feature subspace.

For convenience, we define the matrix $G = MM^TV$ as shorthand. (We will see that $G$ maps
$\omega$ to the movement in agents' feature vectors it incentivizes.)

Agents incur a cost $C(a)$ based on the action they chose.
Throughout this work this cost is quadratic 
$C(a) = \frac{1}{2} \|a\|_2^2$. This corresponds to a setting with
increasing costs to taking any particular action. 

Importantly, we assume that agents will best-respond to the published decision rule by choosing whichever action $a(\omega)$ maximizes
their utility, defined as their received decision minus incurred action cost:\footnote{Note that this means that all strategic agents will, for a given decision rule $\omega$, choose the same gaming action $a(\omega)$.
}
\begin{equation}
    a(\omega) = \arg\max_{a' \in \R^k} \left [ \langle \omega, V(x + Ma')\rangle - \frac{1}{2}\|a'\|^2 \right ]
\end{equation}
However, to take into account the possibility that not all agents will in practice
study the decision rule to figure out the action that will maximize their utility, we further assume that only a
$p$ fraction of agents game the decision rule, while a $1-p$ fraction remain at
their initial feature vector.

Now, the interaction between agents and the decision-maker proceeds in a series of
rounds, where a single round \textit{t} consists of the sequence described in the following figure:

\begin{figure}[ht]\label{fig:setting_sequence}\fbox{
\parbox{\columnwidth}{
For round $t \in \{1,\dots,r\}$:
\begin{enumerate}
    \item The decision-maker publishes a new decision rule $\omega_t$. 
    \item A new set of $n$ agents arrives: $\{x \sim P\}_n$. \\
    Each agent games w.r.t. $\omega_t$; i.e. $x_g \leftarrow x + Ma(\omega_t)$.
    \item The decision-maker observes the post-gaming visible features $V x_g$ for each agent.\\ 
    Agents receive decisions $\omega_t^TV x_g$.
    \item The decision-maker observes the outcome $y \sim {\omega^*}^T x_g + \eta$ for each agent.
\end{enumerate}}}
\end{figure}
In general, we will assume that the decision-maker cares more about minimizing the number of rounds required for an algorithm than the number of agent samples collected in each round.

We now turn to the three objectives that decision-makers may wish to optimize.


    \begin{objective}\label{obj:improvement}
    The \emph{agent outcomes} objective is the average outcome over the agent
    population after gaming:
    \begin{equation}
    \label{eq:improvement}
        \AO(\omega) := \E[x \sim P, \eta]{\langle \omega^*, x + Ma(\omega)\rangle + \eta}
    \end{equation}
    \end{objective}
    In subsequent discussion we will restrict $\omega$ to the unit $\ell_2$-ball, as arbitrarily high outcomes could be produced if $\omega$ were unbounded.
    
    An example of a decision-maker who cares about agent outcomes is a teacher formulating a test for their students---they may
    care more about incentivizing the students to learn the material than about accurately stratifying students based on their knowledge of the
    material.
    
\begin{objective}\label{obj:accuracy}
    \emph{Prediction risk} captures how close the output of the model is to the true
    outcome. It is measured in terms of expected squared error:
    \begin{multline}
    \label{eq:accuracy}
        Risk(\omega) = \mathbb{E}_{x \sim P, \eta} \big[\big (\langle \omega^*,x + Ma(\omega)\rangle \\+ \eta - \langle\omega, V(x +Ma(\omega)) \rangle \big)^2 \big]
    \end{multline}
    \end{objective}

    A decision-maker cares about minimizing prediction risk if they want the scores they
    assign to individuals to be as predictive of their true outcomes as
    possible. For example, insurers' profitability is contingent on neither
    over- nor under-estimating client risk.

    In the realizable linear setting, there is a natural third objective:
    \begin{objective}\label{obj:causal}
    \emph{Parameter estimation} error measures how close the decision rule's
    coefficients are to the visible coefficients of the underlying linear model:
    \begin{equation}\label{eq:causal}
        \|V(\omega - \omega^*)\|_2
    \end{equation}
    \end{objective}
    
    A decision-maker might care about estimating parameters accurately if they want their decision rule to be robust to unpredictable changes in the feature distribution, or if knowledge of the causal effects of the features could help inform the design of other interventions.
    
    Below, we show that these objectives may be mutually non-satisfiable. A natural question is whether we can optimize a weighted combination of these objectives. In Section \ref{sec:risk}, we outline an algorithm for optimizing a weighted combination of prediction risk and agent outcomes. Our parameter recovery algorithm will only work in the fully-visible ($V=I$) case; in this case, all three objectives are jointly satisfied by $\omega = \omega^*$, though each objective implies a different type of approximation error and thus requires a different algorithm.

    \subsection{Illustrative example}\label{sec:example}
\begin{figure}
    \centering
    \includegraphics[width=\linewidth, trim=0 50 0 0, clip]{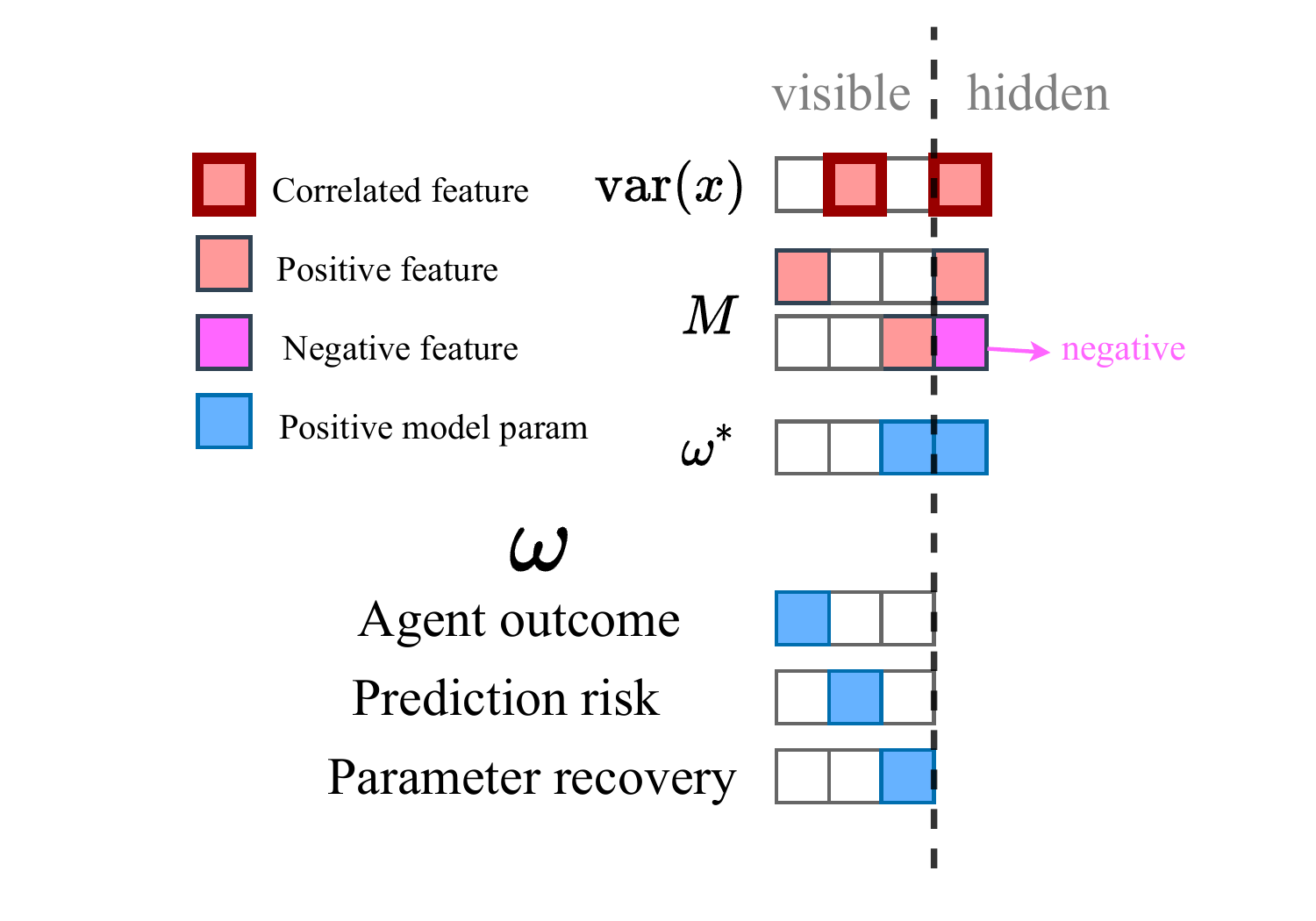}
    \caption{A toy example in which the objectives are mutually non-satisfiable. Each $\omega$ optimizes a different objective.}
    \label{fig:conflicting_objectives}
\end{figure}
    
    To illustrate the setting, and demonstrate that in certain cases these objectives are mutually non-satisfiable, we provide a toy scenario, visualized in Figure \ref{fig:conflicting_objectives}. Imagine a car insurer that predicts customers' expected accident costs given access to three features: (1) whether the customer owns their own car, (2) whether they drive a minivan, and (3) whether they have a motorcycle license. There is a single hidden, unmeasured feature: (4) how defensive a driver they are.
    Let $\omega^* = (0, 0, 1, 1)$, i.e. of these features only knowing how to drive a motorcycle and being a defensive driver actually reduce the expected cost of accidents. Let the initial data distribution have ``driving a minivan'' correlate with defensive driving (because minivan drivers are often parents worried about their kids). Let the first effort conversion column $M_1 = (1, 0, 0, 2)$ represent the action of purchasing a new car, which also leads the customer to drive more defensively to protect their investment. Let the second action-effect column $M_2=(0, 0, 1, -2)$ be the action of learning to ride a motorcycle, which slightly improves the likelihood of safe driving by conferring upon the customer more understanding how motorcyclists on the road will react, while making the customer themselves substantially more thrill-seeking and thus reckless in their driving and prone to accidents.
    
    How should the car insurer choose a decision rule to maximize each objective? 
    If the rule rewards customers who own their own car (1), this will incentivize customers to purchase new cars and thus become more defensive (good for agent outcomes), but will cause the decision-maker to be inaccurate on the $(1-p)$-fraction of non-gaming agents who already had their own cars and no more likely to avoid accidents than without this decision rule (bad for prediction risk), and owning a car does not truly itself reduce expected accident costs (bad for parameter estimation).
    Minivan-driving (2) may be a useful feature for prediction risk because of its correlation with defensive driving, but anyone buying a minivan specifically to reduce insurance payments will not be a safer driver (unhelpful for agent outcomes) nor does minivan ownership truly cause lower accident costs (bad for parameter estimation).
    Finally, if the decision rule rewards customers who have a motorcycle license (3), this does reflect the fact that possessing a motorcycle license itself does reduce a driver's expected accident cost (good for parameter estimation), but an agent going through the process of acquiring a motorcycle license will do more harm than good to their overall likelihood of an accident due to the action's side effects of also making them a less defensive driver (bad for agent outcomes), and rewarding this feature in the decision rule will lead to poor predictions as it is negatively correlated with expected accident cost once the agents have acted (bad for prediction risk).
    
    The meta-point is that when some causal features are hidden from the decision-maker, there may be a tradeoff between the agents' outcomes, the decision-rule's predictiveness, and the correctness of parameters recovered by the regression. In the rest of the paper, we will demonstrate algorithms for finding the optimal decision rules for each objective, and discuss prospects for optimizing them jointly.

\section{Agent Outcome Maximization}
\label{sec:outcome}
In this section we propose an algorithm for choosing a decision rule that
will incentivize agents to choose actions that
(approximately) maximally increase their outcomes. Throughout this section, we will assume that without loss of generality $p=1$ and $|| \omega^\star|| = 1$. If only a subset of agents are strategic ($p<1$), the non-strategic agents' outcomes cannot be affected and can thus be safely ignored.

In our car insurance example, this means
choosing a decision rule that causes drivers to behave the most safely,
regardless of whether the decision rule accurately predicts accident
probability or whether it recovers the true parameters.

Let $\omega_{mao}$ be the decision rule that maximizes agent outcomes:
\begin{equation}
    \omega_{mao} := \argmax_{\omega \in \R^d, \|\omega\|_2 \leq 1} \AO(\omega)
\end{equation}

\begin{theorem}\label{thm:improve}
Suppose the feature vector second moment matrix $\Sigma$ has largest eigenvalue bounded above by $
\lambda_{max}$, and suppose the outcome noise $\eta$ is 1-subgaussian. Then
Algorithm~\ref{alg:improvement} estimates a parameter vector $\hat\omega$ in
$d+1$ rounds with $O(\lambda_{max} \epsilon^{-2} d^2)$ samples in each round such that w.h.p.
\[ \AO(\hat\omega) \geq \AO(\omega_{mao}) - \epsilon. \]
\end{theorem}
\begin{algorithm}
   \caption{Agent Outcome Maximization}
   \label{alg:improvement}
\begin{algorithmic}
   \STATE {\bfseries Input:} $\lambda_{max}, d, \epsilon$ 
   \STATE Let $n = 100 \epsilon^{-1} \lambda_{max}d$
   \STATE Let $\{\omega_j\}_{i=1}^d$ be an orthonormal basis for $\R^d$
   \STATE Sample $(x_1,y_1) \ldots (x_n,y_n)$ with $\omega = 0$.
   \STATE Let $\hat \mu = \frac{1}{n} \sum_{j=1}^n y_j$
   \FOR{$i=1$ {\bfseries to} $d$}
    \STATE Sample $(x_1,y_1) \ldots (x_n,y_n)$ with $\omega = \omega_i$
    \STATE Let $\hat\nu_i = \frac{1}{n} \sum_{j=1}^n y_j - \hat\mu$
   \ENDFOR
   \STATE Let $\hat\nu = (\hat\omega_{mao}^{(1)},\dots,\hat\omega_{mao}^{(d)})^T$
   \STATE Let $\hat\omega = \hat\nu/\|\hat\nu\|$
   \STATE Return {$\hat\omega$}
\end{algorithmic}
\end{algorithm}
\begin{proofidea}
First we note that it is straightforward to compute the action that each agent
will take. Each agent maximizes $\omega^T V (x + Ma) - \frac 1 2 \|a\|^2$ over $a
\in \R^m$. Note that $\nabla_a(\omega^T V Ma - \frac 1 2 \|a\|^2)= M^TV\omega  -
a$. Thus,
\begin{align*}
    &\argmax\limits_{a} \omega^T V (x + Ma) - \frac 1 2 \|a\|^2\\
    &=\argmax\limits_{a} \omega^T V Ma - \frac 1 2 \|a\|^2\\
    &= M^T V \omega
\end{align*}
That is, every agent chooses an action such that $x_g = x + MM^TV\omega = x + G\omega$ (recall we have defined $G := MM^T V$ for notational compactness).
This means that if the decision-maker publishes $\omega$, the resulting
expected agent outcome is $\AO(\omega) = \E[x \sim P]{{\omega^*}^T x + {\omega^*}^T G\omega)}$. Hence,
\[ \omega_{mao} = \frac{G^T \omega^*}{\|G^T \omega^*\|_2} \]

In the first round of the algorithm, we set $\omega = 0$ and obtain an empirical estimate of $\AO(0) = \E[x \sim P]{{\omega^*}^T x}$.\footnote{Alternatively, the decision-maker could run this algorithm but with all parameter vectors shifted by some $\omega$.} We then select an orthonormal basis $\{\omega_i\}_{i=1}^d$ for $\R^d$. In each subsequent round, we publish an $\omega_i$ and obtain an estimate of $\AO(\omega_i)$. Subtracting the estimate of $\AO(0)$ yields an estimate of $\E[x \sim P]{{\omega^*}^T G \omega_i}$, which is a linear measurement of $G^T \omega^*$ along the direction $\omega_i$. Combining these linear measurements, we can reconstruct an estimate of $G^T \omega^*$. The number of samples per round is chosen to ensure that the estimate of $G^T \omega^*$ is within $\ell_2$-distance at most $\epsilon/2$ of the truth. We set $\hat\omega$ to be this estimate scaled to have norm 1. A simple argument allows us to conclude that $\AO(\hat\omega)$ is within an additive error of $\epsilon$ of $\AO(\omega_{mao})$. We
leave a complete proof to the appendix.
\end{proofidea}

This algorithm has several desirable characteristics. First, the
decision-maker who implements the algorithm does not need to have any knowledge
of $M$ or even of the number of hidden features $d' - d$. Second, the algorithm
is non-adaptive, in that the published decision rule in each round does not
depend on the results of previous rounds. Hence, the algorithm can be
parallelized by simultaneously applying $d$ separate decision
rules to $d$ separate subsets of the agent population and simultaneously
observing the results. Finally, by using decision rules as causal interventions, this procedure resolves the challenge associated
with the hardness result of \cite{miller2020}.


\section{Prediction Risk Minimization}

\label{sec:risk}
Low prediction risk is important in any setting where the decision-maker wishes
for a decision to accurately match the eventual outcome. For example, consider an insurer
who wishes to price car insurance exactly in line with drivers' expected costs
of accident reimbursements. Pricing too low would make them unprofitable, and pricing too high would allow a competitor to undercut them.

Specifically, we want to minimize expected squared error when predicting the
true outcomes of agents, a $p$-fraction of whom have gamed with respect to $\omega$:
\begin{equation}\label{eqn:acc}
    \begin{split}
    Risk(\omega) &= \mathbb{E}_{x \sim P, \eta} \Bigg [
        (1-p) \left (\omega^TVx - (\omega^*)^T x \right )^2 \\
      &+ p\left (\omega^T V x_g - \left ({\omega^*}^T x_g + \eta \right )\right )^2 \Bigg ]
    \end{split}
\end{equation}
We begin by noting a useful decomposition of accuracy in a generalized linear setting. For this result, all that we assume about agents' actions is that agents' feature vectors and actions are drawn from some joint distribution $(x,a) \sim D$ such that the action effects are uncorrelated with the features: $\E[(x,a) \sim D]{(Ma)x^T} = 0$.
\begin{lemma}
\label{lem:decomp}
Let $\omega$ be a decision rule and let $a$ be the action taken by agents in
response to $\omega$. Suppose that the distributions of $Ma$ and $x$ satisfy $\E[x,a]{(Ma)x^T} = 0$. Then the expected squared error of a decision rule $\omega$ on the gamed
distribution can be decomposed as the sum of the following two positive terms
(plus a constant offset $c$):
\begin{enumerate}
    \item The risk of $\omega$ on the initial distribution
    \item The expected squared error of $\omega$ in predicting the impact (on agents'
    outcomes) of gaming.
\end{enumerate}
That is,
    \begin{equation}
        \label{eq:decomp}
        \begin{split}
        Risk(\omega) &= \E[x]{\left( (V\omega - \omega^*)^Tx\right)^2} \\
        &+ \E[a]{\left( (V\omega - \omega^*)^T(Ma)\right )^2} 
        + c
        \end{split}
    \end{equation}
\end{lemma}

The proof appears in the appendix.

This decomposition illustrates that minimizing prediction
risk requires balancing two competing phenomena. First, one must minimize the risk associated with
the original (ungamed) agent distribution by rewarding 
features that are \emph{correlated with outcome} in the original data. Second one must
minimize error in predicting the effect of agents' gaming on their
outcomes by rewarding features in accordance with \emph{the true
change in outcome}. 
The relative importance of these two phenomena depends on  $p$, the fraction of agents who game.

Unfortunately, minimizing this objective is not straightforward. Even with just squared action cost (with actions $a(\omega)$ linear in
$\omega$), the objective becomes a non-convex quartic.
However, we will show that in cases where the naive gaming-free predictor \emph{overestimates} the impact of gaming, this quartic can be minimized efficiently.

\begin{algorithm}
  \caption{Relaxed Prediction Risk Oracle}
  \label{alg:risk}
\begin{algorithmic}
  \STATE {\bfseries Input:} $\omega, n$ 
  \STATE Let $P_{\omega}$ be the distribution of agent features and labels $(x, y)$ drawn when agents face decision rule $\omega$.
  \STATE Let $Y(S) \coloneqq \frac{1}{\|S\|} \sum_{(x, y) \in S} y$.
  \STATE Let $\tilde{Y}_\omega (S) \coloneqq \frac{1}{\|S\|} \sum_{(x, y) \in S} (V\omega)^T x$.

  
  \STATE Collect samples $D_i = \left \{(x, y) \sim P_{\omega_i} \right \}_{n}$.
  
  \IF{$\tilde{Y}_{\omega}(D_i) > Y(D_i)$}
    \STATE Collect new samples $D'_i =\left \{(x, y) \sim P_{\omega_i} \right \}_n$.

  \RETURN $\frac{1}{n}\sum_{(x, y) \in D'_i} ((V\omega_i)^Tx - y)^2$
  \ELSE
  \STATE Collect\footnote{Note that, if we know the number of rounds $k$, we can simply deploy the $\vec{0}$ once initially and collect $kn$ samples, halving the total round complexity.} samples $D^*_i = \left \{(x, y) \sim P_{\vec{0}} \right \}_n$.
  \RETURN $\frac{1}{n}\sum_{(x, y) \in D^*_i} ((V\omega_i)^Tx - y)^2$
  \ENDIF
\end{algorithmic}
\end{algorithm}

\begin{remark}
    \label{remark:risk}
    Let $\omega_{\text0}$ be the decision rule that minimizes the
    prediction risk without gaming and let agent action costs be $C(a) = \frac{1}{2}\|a\|_2^2$.
    If $\omega_{\text0}$ \emph{overestimates} the change in agent outcomes as
    a result of the agents' gaming, then we can find
    an approximate prediction-risk-minimizing decision rule in $k =O(\mathrm{poly}(d))$ rounds by using a derivative-free optimization procedure on the convex relaxation implemented in Algorithm \ref{alg:risk}.
\end{remark}
\begin{proofidea}
To find a decision rule $\omega$ that minimizes predictive risk, we first need to define prediction risk as a function of $\omega$.
As shown in Lemma \ref{lem:decomp}, the prediction risk $Risk(\omega)$ consists of two terms:
the classic gaming-free prediction risk (a quadratic),
and the error in estimating the effect of gaming on the outcome (the ``gaming error''). In the quadratic action cost case, this can be written out as
$\left ( (V\omega - \omega^*)^T(MM^TV\omega)\right )^2$. This overall objective is a high-dimensional quartic, with large nonconvex regions. Instead of optimizing this sum of a convex function and nonconvex function, we optimize the convex function plus a convex relaxation of the nonconvex function. Since the minimum of the original function is in a region where the relaxation matches the original, this allows us to find the optimum using standard optimization machinery. 

The trick is to observe that of the two terms composing the prediction risk, the prediction risk before gaming is always convex (it is quadratic), and the ``gaming error'' is a regular quartic that is convex in a large region. Specifically, there exists an ellipse in decision rule space separating the convex from the (partially) non-convex regions of the ``gaming error'' function, where the value of this ``gaming error'' on the ellipse is exactly $0$. To see why, note that there is a possible rotation and translation of $\omega$ into a vector $z$ such that the quantity inside the square, $(V\omega - \omega^*)^TMM^TV\omega$, can be rewritten as $z_1^2 + z_2^2 + \ldots + z_d^2 -c$, where $c>0$. The zeros of this function, and thus of the ``gaming error'' (which is its square), form an ellipse.
This ellipse corresponds to the set of points where the decision rule $\omega$ perfectly predicts the effect on outcomes of gaming induced by agents responding to this same decision rule, meaning $\omega^{*T}Ma - \omega^TVMa = 0$. In the interior of this ellipse, where the decision rule underestimates the true effect of gaming on agent outcomes ($\omega^{*T}Ma > \omega^TVMa$), the quartic is not convex. On the other hand, on the area outside of this ellipse, where $\omega^{*T}Ma < \omega^TVMa$, it is always convex.

$\omega_0$ minimizes the prediction risk on ungamed data, and is thus the minimum of the quadratic first term.
We are given that $\omega_0$ overestimates the effect of gaming on the outcome $\omega_0^{*T}Ma < \omega_0^TVMa$ and is thus outside the ellipse.\footnote{Note that if $\omega_0^{*T}Ma = \omega_0^TVMa$ exactly, then $\omega_0$ is a global optimum and we are done.}
We will now show that a global minimum lies outside the ellipse. Assume, for contradiction, that all global minima lie inside the ellipse. Pick any minimum $\omega_{min}$. Then there must be a point $\omega'$ on the line between $\omega_0$ and $\omega_{min}$ that lies on the ellipse. All points on the ellipse are minima (zeros) for the ``gaming error'' quartic, so the second component of the predictive risk is weakly smaller for $\omega'$ than for $\omega_{min}$. But the value of the quadratic is also smaller for $\omega'$ since it is strictly closer to the minimum of the quadratic $\omega_0$. Thus $Risk(\omega') < Risk(\omega_{min})$, which is a contradiction. This means that the objective $Risk(\omega)$ is convex in a neighborhood around its global minimum, which guarantees that optimizing this relaxation of the original objective yields the correct result. 

The remaining concern is what to do if we fall into the ellipse, and thus potentially encounter the non-convexity of the objective. 
We solve this by, in this region, replacing the overall prediction risk objective $Risk(\omega)$ with the no-gaming prediction risk objective (on data sampled when $\omega = \vec{0}$). Geometrically, the function on this region becomes a quadratic. The values of this quadratic perfectly match the values of the quartic at the boundary, so the new piecewise function is convex and continuous everywhere. In practice, we empirically check whether the decision rule on average underestimated the true outcome. If so, we substitute the prediction risk with the classic gaming-free risk.

We can now give this objective oracle to a derivative-free convex optimization procedure, which will find a decision rule with value $\epsilon$-close to the global prediction-risk-minimizing decision rule in a polynomial (in $d$, $1 / \epsilon$) number of rounds and samples.
\end{proofidea}
This raises an interesting observation: in our scenario it is easier to recover
from an initial over-estimate of the effect of agents' gaming on the outcome (by
reducing the weights on over-estimated features)
than it is to recover from an under-estimate (which requires increasing the
weight of each feature by an unknown amount).

\begin{remark}
    The procedure described in Remark \ref{remark:risk} can also be used to
    minimize a weighted sum of the outcome-maximization and
    prediction-risk-minimization objectives.
\end{remark}
This follows from the fact that the outcome-maximization
objective is linear in $\omega$, and therefore adding it to the prediction-risk
objective preserves the convexity/concavity of each of the different regions of
the objective. Thus, if a credit scorer wishes to find the loan decision rule that maximizes a weighted sum of their
 accuracy at assigning loans, and the fraction of their customers who successfully repay (according to some weighting), this provides a method for doing so under certain initial conditions.

\section{Parameter Estimation}
\label{sec:parameter_estimation}

Finally, we provide an algorithm for estimating the causal outcome-generating parameters $\omega^*$, specifically in the case where the features are fully visible ($V=I$).\footnote{For simplicity, we also assume $p=1$, though any lower fraction of gaming agents can be accomodated by scaling the samples per round.}
\begin{restatable}{theorem}{causal}
\label{thm:causal}
(Informal) Given $V = I$ (all dimensions are visible) and $\Sigma + \lambda MM^T$ is full rank for some $\lambda$ (that is, there exist actions that will allow change in the full feature space), we can estimate $\omega^*$ to arbitrary precision. 
We do so by computing an $\omega$ that results in more informative samples, and then gathering samples under that $\omega$. The procedure requires $\widetilde O(d)$ rounds. See the appendix for details.) 
%
\end{restatable}
The algorithm that achieves this result consists of the following steps:
\begin{enumerate}
    \item Estimate the covariance of the initial agent feature distribution before strategic behavior $\Sigma$ by initially not disclosing any decision rule to the agents, and observing their features.
    \item Estimate parameters of the Gramian of the action matrix $G=MM^T$ by incentivizing agents to vary each feature sequentially.
    \item Use this information to learn the decision function $\omega$ which will yield the most informative samples in identifying $\omega^*$, via convex optimization.
    \item Use the new, more informative samples in order to run OLS to compute an estimate of the causally precise regression parameters $\omega^*$.  
\end{enumerate}
At its core, this can be understood as running OLS after acquiring a better dataset via the smartest choice of $\omega$ (which is, perhaps surprisingly, unique). Whereas the convergence of OLS without gaming would be controlled by the minimum eigenvalue of the second moment matrix $\Sigma$, convergence of our method is governed by the minimum eigenvalue of post-gaming second-moment matrix: 
$$ \mathbb E[(x + G \omega) (x + G \omega)^T] = \Sigma + 2 \mu \omega^T G^T + G \omega \omega^T G^T$$
Our method learns a value of $\omega$ that results in the above matrix having a larger minimum eigenvalue, improving the convergence rate of OLS. The proof and complete algorithm description is left to the appendix.

\section{Discussion}
In this work, we have introduced a linear setting and techniques for analyzing
decision-making about strategic agents capable of changing their outcomes.
We provide algorithms for leveraging agents' behaviors to maximize agent
outcomes, minimize prediction risk, and recover the true parameter vector governing the outcomes.
 Our results suggest that in certain settings, decision-makers would benefit by not just being passively transparent about their decision rules, but by actively informing strategic agents.

While these algorithms eventually yield more desirable decision-making
functions, they substantially reduce the decision-maker's accuracy in the short
term while exploration is occurring. Regret-based analysis of causal strategic learning is a potential avenue for future work.
In general, these procedures make the most sense in scenarios with a fairly
small period of delay between decision and outcome (e.g. predicting short-term
creditworthiness rather than long-term professional success), as at each new
round the decision-maker must wait to receive the
samples gamed according to the new rule.

Our methods also rely upon the assumption that new agents appear every round. If there are persistent stateful agents, as in many real-world repeated decision-making settings, different techniques may be required.

\section*{Acknowledgements}

The authors would like to thank 
        Suhas Vijaykumar,
        Cynthia Dwork,
        Christina Ilvento,
        Anying Li,
        Pragya Sur,
        Shafi Goldwasser,
        Zachary Lipton,
        Hansa Srinivasan,
        Preetum Nakkiran,
        Thibaut Horel,
        Fan Yao,
        and our anonymous reviewers
        for their helpful advice and comments.
        Ben Edelman was partially supported by NSF Grant CCF-15-09178.
        Brian Axelrod was partially supported by NSF Fellowship Grant DGE-1656518, NSF Grant 1813049, and NSF Award IIS-1908774.

\bibliography{bibliography}
\bibliographystyle{icml2020_style_final/icml2020}

\onecolumn

\section{Appendix}

\subsection{Agent Outcomes}

\begin{proof}[Proof of Theorem~1]
Let's walk through the steps of the algorithm, bounding the error that accumulates along the way.

In the first round we set $\omega = 0$ in order to obtain an estimate for
$E[{\omega^*}^T x]$.

Since $\omega^*$ is a unit vector, the variance of
${\omega^*}^T x$ is at most $\lambda_{max}$ plus a constant (from the
$1-$subgaussian noise).

By Chebyshev's inequality, this means that $O(\lambda_{max} \epsilon^{-2} d^2 )$ samples suffice for the
empirical estimator of $E[{\omega^*}^T x]$ to have no more than $\frac \epsilon
{4d}$ error with failure probability $\Omega (\frac 1 {2d})$. We call the output
of this estimator $\hat \mu$ and let $\hat \mu_d$ be the r-dimensional vector
with $\hat \mu$ in every coordinate. 

Now we choose $\omega_1....\omega_d$ that form an orthonormal basis of the image
of the diagonal matrix $V$. For each $\omega$ we observe the reward
${\omega^*}^T (x + G\omega) + \eta$, subtract out $\hat \mu$, and plug it into
the empirical mean estimator. For each $\omega_i$, let $\hat \nu_i$ be the
resulting coefficient. After $O(\epsilon^{-1} d \lambda_{max})$ samples, each coefficient has
at most $\frac \epsilon {4d}$ error with failure probability at most $\frac 1
{2d}$. Since we have computed $d+1$ estimators, each one with failure
probability at most $\frac 1 {2d}$, a union bound gives us a total failure
probability that is sub-constant.

We can now bound the total squared $\ell_2$ error between said coefficients and
$G^T \omega^*$ in the $\omega_1...\omega_d$ basis (noting that the choice of
basis does not affect the magnitude of the error).
We can break up the error into two components using the triangle inequality: the
error due to $\hat \mu_d$ and the error in the subsequent rounds. Each
coordinate of $\hat \mu_d$ has error of magnitude at most $\frac \epsilon {4d}$, so the
total magnitude of the error in $\hat \mu_d$ is at most $\frac \epsilon {4}$.
The same argument applies for the error in the coordinate estimates, leading to
a total $\ell_2$ error of at most $\epsilon/2$.


Recall that $\hat\omega = \hat\nu/\|\hat\nu\|$. Let $\nu := G^T \omega^*$. We can now bound the gap between the agent outcomes incentivized by $\hat\omega$ and by $\omega_{imp} = \nu/\nu$:
\begin{align}
\AO(\omega_{imp}) - \AO(\hat\omega) &= \nu^T \frac{\nu}{\|\nu\|} - \nu^T \frac{\hat\nu}{\|\hat\nu\|} \\
&= \|\nu\| - \nu^T \frac{\hat\nu}{\|\hat\nu\|} \\
&\leq \|\nu\| - \frac{\|\nu\|(\|\nu\|-\epsilon/2)}{\|\nu\|+\epsilon/2} \\
&= \frac{\|\nu\|\epsilon}{\|\nu\|+\epsilon/2} \leq \epsilon
\end{align}
\end{proof}

\subsection{Prediction Risk}

\begin{proof}[Proof of Lemma~1]

\begin{align*}
    Risk(\omega) &= \E[x,a]{\left (\omega^T V \left (x + Ma\right ) - {\omega^*}^T\left (x+Ma \right ) \right )^2} \\
    =& \E[x,a]{\left ( \left (\omega^T Vx - {\omega^*}^T x \right) + \left (\omega^T VMa - {\omega^*}^T Ma\right ) \right )^2} \\
    =& \E[x,a]{\left (\omega^T Vx - {\omega^*}^T x\right)^2} + \E[x,a]{\left(V\omega - \omega^*\right)^T x (Ma)^T \left(V\omega - \omega^*\right)} + \E[x,a]{\left (\omega^T VMa - {\omega^*}^T Ma\right )^2} \\
    =& \E[x]{\left (\omega^T Vx - {\omega^*}^T x\right)^2} + \E[a]{\left (\omega^T VMa - {\omega^*}^T Ma\right )^2}
\end{align*}
where the last line follows because $Ma$ and $x$ are uncorrelated.
\end{proof}

\subsection{Parameter Estimation}
\label{app:causal}
In this section we describe how we recover $\hat\omega_{opt}$ in $L^2$-distance when there exists an $\omega$ such that $\Sigma + G \omega$ is full rank. Before we proceed we make a couple of observations. When there is no way to make the above matrix full rank, we cannot hope to recover the optimal $\hat\omega_{opt}$. If there is no natural variation in e.g. the last two features, and furthermore no agent can act along those features, it is not possible to disentangle their potential effects on the outcome. This also suggests that the parameter recovery is a more substantive demand for the decision maker than the standard linear regression setting. To discover this additional information, the decision maker can incentivize the agents to take actions that help the decision-maker recover the true outcome-governing parameters.

This motivates the algorithm we present in this section. It operates in two stages. First, it recovers the information necessary in order to to identify the decision rule which will provide the most informative agent samples after those agents have gamed. Second, it collects data while incentivizing this action. Finally, it computes an estimate of $\hat\omega_{opt}$ using the collected data. We present the complete procedure in Algorithm \ref{alg:causal}. 

\begin{algorithm*}
  \caption{Recovering the Causal Model} 
  \label{alg:causal}
  \begin{algorithmic}[1]
  \STATE Let $k_1 = \lambda_{max}(G^TG)$ and $k_2 = || \Sigma ||^2$
  \STATE Let $\kappa_{min} = \lambda_{min} (\Sigma)$
  \STATE Choose an $\epsilon > 0$
  \STATE Let $n_1 = O(\max (\frac{d k_1}{\kappa_{min}}, \frac{d^2 k_2}{\kappa_{min}}) )$
  \STATE Collect samples $x_1,\ldots,x_{n_1}$
  \STATE Let $\hat \mu = \frac{1}{n_1}\sum x_i$
  \STATE Let $\hat \Sigma = \frac{1}{n_1}\sum x_ix_i^T$
  \STATE Let $n_2 = O( \max({d^2 ||\hat\mu||^2 \mathrm{tr}(\Sigma), d^3 ||G||^2 \mathrm{tr}(\Sigma)  }))$
  \FOR{ $i = 1...d$}
  \STATE $\omega = e_i$
  \STATE Sample $x_1, \ldots, x^i_{n_2}$ and subtract $\hat \mu$ from each one. 
  \STATE Let $\hat G_i = \frac{1}{n_2} \sum\limits_{j = 1}^{n_2} x_j$ 
  \ENDFOR
  \STATE Let $\hat\omega_{opt} = \argmin\limits_{\omega} \hat \Sigma + 2 \mu \omega^T \hat G^T +  \hat G \omega \omega^T \hat G ^T $
  \STATE Let $n_3 = O(\frac{d}{\epsilon \kappa_{min}})$ 
  \STATE Sample $x_1,\ldots, x_{n_3}$ with $\omega = \hat\omega_{opt}$. 
  \STATE Return the output of OLS on $x_1,\ldots, x_{n_3}$
  \end{algorithmic}
\end{algorithm*}

The procedure in Algorithm \ref{alg:causal} can be summarized as follows: 
\begin{enumerate}
    \item Estimate the first and second moments of the distribution of agents' features.
    \item Estimate the Gramian of the action matrix $G$.
    \item Compute the most informative choice of $\omega$.
    \item Collect samples under the most informative $\omega$ and then return the output of OLS.
\end{enumerate}

Before we proceed to the proof of correctness of Algorithm \ref{alg:causal}, let us build some intuition for why this procedure of choosing a single $\omega$ and collecting samples under said $\omega$ makes sense. As we show later, the convergence of OLS for linear regression can be controlled by the minimum eigenvalue of the second moment matrix of the samples. Our algorithm finds the value of $\omega$ that, after agents game, maximizes this minimum eigenvalue in expectation. It turns out the minimum eigenvalue of the expected second moment matrix of post-gaming samples is convex with respect to the choice of $\omega$. The convexity of the objective suggests that a priori, when choosing $\omega$s to obtain informative samples, the optimal strategy is choose a single specific $\omega$. 

The main difficulty in the rest of the algorithm is achieving the necessary precision in the estimation to be able to set up the above optimization problem to identify such an $\omega$. 

\textbf{Theorem 3. }\textit{
When $V = I$, the output of Algorithm \ref{alg:causal} run with parameter $\epsilon$ satisfies $|| \omega - \omega^*|| \le \epsilon$ with probability greater than $\frac 2 3$.}

The proof of Theorem 3 relies on several lemmas. First we bound the $L_2$ error of OLS as a function of the empirical second moment matrix in Lemma \ref{lem:olsell2}. Note that the usual bound for the convergence of OLS is distribution dependent. That is, the expected error is small. 

\begin{lemma}\label{lem:olsell2}
Assume $V = I$. Consider samples $x_1,\ldots, x_n$ and $y_i = {\hat\omega_{opt}}^T x_i + \eta_i$. Let $\omega$ be the output of OLS $(x_i, y_i)$. Then 
$$ \mathbb E_{\eta}\left [ || \omega - \hat\omega_{opt}|| ^2 \right] \le \frac{d}{n \kappa_{min}}$$
\end{lemma}
The above proof is elementary and a slight modification of the standard textbook proof (see for example, \cite{liangstat}).

The proof also requires that the optimization to choose the optimal $\omega$ is convex. 
\begin{lemma}\label{lem:convex}
The minimum eigenvalue of the following matrix is convex with respect to $\omega$ for any values of $x,G$. 
$$ \sum\limits_i (x_i + \hat G \omega)(x_i + \hat G \omega)^T$$ 

Furthermore, when the following conditions are true, then the minimum eigenvalue of the above is within a constant factor of the optimal value. 

$\mathbb E[(x + G \omega)(x + \hat G \omega)^T ]$. 

\begin{itemize}
    \item $|| \hat \Sigma - \Sigma ||^2 \le \epsilon$
    \item $|| \mu - \hat \mu || ^2 \le \frac{\lambda_{max}(G^T G) \epsilon }{d}$
    \item $ ||\hat G - G||^2 \le \frac{\epsilon}{d || \mu ||^2}$
    \item $ || \hat G - G ||^2 \le \frac{\epsilon}{d^2 || G ||^2}$
\end{itemize}

Finally, the above holds true even for an $\omega$ with distance at most $O(\frac{1}{poly(d)})$ from the optimum.
\end{lemma}

Finally, we use a minor lemma for recover of a random vector via the empirical mean estimator. Note that we treat the matrix $G$ as a vector. 
\begin{lemma}\label{lem:grec}
Assume $V = I$. Let $g_1^i,\ldots, g_n^i$ be drawn from the distribution $G_i + \xi$ and $\hat G$ be the empirical mean estimator computed from said $g_j^i$'s. 
Let $\Sigma$ be the expected second moment matrix of the $\xi$s. 
Then $$ \mathbb E_\xi ||G - \hat G ||^2 \le \frac{d^2 \mathrm{tr}(\Sigma)}{n}$$
\end{lemma}

We proceed with the proof of Theorem 3 below. 

\begin{proof}
The first step of the algorithm is for recovering an estimate of $\Sigma$ and $\mu$.
Note that $n_1$ samples suffice to recover $\hat \Sigma$ and $\hat \mu$ such that:
\begin{itemize}
    \item $|| \hat \Sigma - \Sigma ||^2 \le \epsilon$
    \item $|| \mu - \hat \mu || ^2 \le \frac{\lambda_{max}(G^T G) \epsilon }{d}$
\end{itemize}

The for loop recovers an estimate of $G$. Via Lemma \ref{lem:grec}, the samples suffice to ensure that the following two conditions hold: 
\begin{itemize}
    \item $ ||\hat G - G||^2 \le \frac{\epsilon}{d || \mu ||^2}$
    \item $ || \hat G - G ||^2 \le \frac{\epsilon}{d^2 || G ||^2}$
\end{itemize}

Then the algorithm computes an estimate of the optimal $\omega$. Via Lemma \ref{lem:convex}, we have that the optimum guarantees the minimum eigenvalue of an approximate solution will be within a constant factor of the optimum. 

This $\omega$ guarantees that $n_3$ samples suffice to ensure the recover of $\omega^*$ within squared $L^2$-distance of $O(\epsilon)$ in expectation. 

Finally the expectations can be used with a Markov inequality to ensure the algorithm succeeds with (arbitrarily high) constant probability. 
\end{proof}

Now we prove the lemmas. 
We begin with Lemma \ref{lem:olsell2}. This proof is a slight modification of the textbook proof for the convergence of OLS. 

\begin{proof}
In this section we derive a bound on the convergence of the least squares estimator when a fixed design matrix $X$ is used. Note this is exactly the case we encounter, since the choice of $\omega$ lets us affect the entries of the design matrix. This is a standard, textbook result and not a main contribution of the paper. 

In order to state the result more formally we have to introduce some notation. The goal of the procedure is to recover $\hat\omega_{opt}$, when given tuples $(x_i, \hat\omega_{opt} x_i + \eta)$ where $\eta$ is 1-subgaussian. We aim to characterize $||\omega - \hat\omega_{opt} || $ where $\omega$ is obtained from ordinary least squares. Let $X$ be the vector with the $x_i$'s in its columns. Let $\kappa_{min}$ be the minimum eigenvalue of $\frac 1 n X^TX$ (the second moment matrix).  

Below all expectations are taken \emph{only} over the random noise. We assume the second moment matrix is full rank.

\begin{align*}
    \mathbb E[|| \omega - \hat\omega_{opt}|| ^2] &\le \mathbb E[ \frac{1}{n\kappa_{min}}  (\omega - \hat\omega_{opt}) X^T X (\omega - \hat\omega_{opt})]\\
    & = \mathbb E[\frac{1}{n\kappa_{min}} || X (\omega - \hat\omega_{opt})|| ^2] \\
    &= \frac{1}{n\kappa_{min}} \mathbb E[|| X(X^TX)^{-1}X^T (X \hat\omega_{opt} + \eta) - X\hat\omega_{opt}|| ^2]\\
    &= \frac{1}{n\kappa_{min}} \mathbb E[|| X(X^TX)^{-1}X^T \eta|| ^2]\\
    &\le \frac{d}{n \kappa_{min}}
\end{align*}

This motivates our procedure for parameter recovery. We do so in a fashion that attempts to maximize $\kappa_{min}$. Note that it is the minimum eigenvalue that determines the convergence rate. This is due to the fact that little variation along a dimension makes it hard to disentangle the features' effect on the outcome via $\hat\omega_{opt}$ from the constant-variance noise $\eta$.  
\end{proof}

Lemma \ref{lem:convex} is somewhat more involved. It is proven in three parts. The first is that the optimization problem is convex. The second is that approximate recovery of $S, \mu,$ and  $G$ suffice for approximately minimizing the original expression. The third is that an approximate solution suffices. 

\begin{proof}
In this section we describe how to choose the value of $\omega$ that maximizes the value of $\kappa_{min}$ for the samples we obtain. 

To do so, we examine the expectation of the second moment matrix and make several observations. Let $\Sigma$ denote the expected second moment matrix of $x$ (i.e. $\mathbb E[xx^T]$. We have: 
$$ \mathbb E[(x + G \omega) (x + G \omega)^T] = \Sigma + 2 \mu \omega^T G^T + G \omega \omega^T G^T$$
\begin{enumerate}
    \item The minimum eigenvalue of the above expression is concave with respect to $\omega$. This follows due to the following: $x + G \omega$ is a linear operator, the minimum eigenvalue of a Gramian matrix $X^T X$ is concave with respect to $X$, and the expectation of a concave function is concave \cite{boyd2004convex}. 
    \item Since the agent attempts to maximize their motion in the $\omega$ direction, we want to ensure that we move them toward toward the direction that maximizes the minimum eigenvalue of $\mathbb E[ (x + G \omega) (x + G \omega)^T]$. 
\end{enumerate}

However, we do not operate with exact knowledge of $G$, etc. It turns out that even approximately solving this optimization problem with estimates for $G, \Sigma, \mu$ suffices for our purposes, as long as the $\omega$ we obtain from our optimization (using the estimates) results in a high value for the minimum eigenvalue of $\mathbb E[(x + G \omega) ( x + G \omega)^T] $. Let $\hat \omega$ be the maximizing argument for the estimated optimization problem and let $\omega$ be the maximizing argument for the original optimization problem. Let $Q$ be the true maximized second moment matrix including gaming, and $\hat Q$ be the maximizing second moment matrix with gaming resulting from replacing the true $\Sigma, \mu, G$ with the estimates. In formal terms, we need to show the minimum eigenvalue of the following is large: $ \mathbb E [(x + G \hat \omega) (x + G \hat \omega)^T] $. 
We note that when $y^T \hat Q y$ is within $\epsilon$ of $y^T Q y$ for all $y$ in the unit ball, the minimum eigenvalues may differ by at most $\epsilon$. 
\begin{align*}
    || y^T \hat Q y - y ^T Q y||^2 &= || y^T (\hat Q - Q) y||^2\\
    &\le \lambda_{max}^2 (\hat Q - Q)(y) || y ||^2\\
    & \le || \hat Q - Q||^2 
\end{align*}

And now we bound the norm of $|| \hat \Sigma - \Sigma ||^2$ assuming the following:
\begin{enumerate}
    \item $|| \hat \Sigma - \Sigma||^2 \le \epsilon $
    \item $|| \mu - \hat \mu ||^2 \le \frac{\lambda_{max}(G^T G) \epsilon}{d}$
    \item $|| \hat G - G ||^2 \le \frac \epsilon {d ||\mu||^2}$
    \item $|| \hat G - G ||^2 \le \frac{\epsilon} {d^2 || G ||^2}$
\end{enumerate}

We work out the bound below. 
\begin{align*}
    || \hat Q - Q ||^2 &= || \Sigma + 2 \mu {\hat\omega_{opt}}^T G^T + G \hat\omega_{opt} {\hat\omega_{opt}}^T G^T - (\hat \Sigma + 2 \hat \mu \omega^T \hat G^T + \hat G  \hat\omega_{opt} {\omega_{opt}}^T \hat G^T)||^2\\
    &\le || \Sigma - \hat \Sigma|| ^2 + 2||  \mu {\hat\omega_{opt}}^T G^T -  \hat \mu {\hat\omega_{opt}}^T \hat G^T|| ^2 + || \ldots|| ^2\\
    & \le \epsilon + 2||  \mu {\hat\omega_{opt}}^T G^T  + {\hat \mu \hat\omega_{opt}}^TG^T- {\hat \mu \hat\omega_{opt}}^TG^T-  \hat \mu {\hat\omega_{opt}}^T \hat G^T|| ^2 + || \ldots|| ^2\\
    &\le \epsilon + d|| \mu - \hat \mu||^2 ||  {\hat\omega_{opt}}^T G ||^2 + || \hat G - G||^2 || \hat \mu \omega^T||^2 + \ldots\\
    &\le \epsilon + \epsilon + || \hat G - G||^2 || \hat \mu \omega^T  +\mu  \omega^T - \mu \omega^T ||^2 + \ldots\\
    &\le \epsilon + \epsilon + || \hat G - G||^2 (|| \hat \mu - \mu||^2 + || \mu {\hat\omega_{opt}}^T||) + \ldots\\
    & \le \epsilon + \epsilon + || \hat G - G||^2 d || \mu||^2 + \ldots\\
    & \le 3 \epsilon + || \hat G  \hat\omega_{opt} {\hat\omega_{opt}}^T \hat G^T -  G  \hat\omega_{opt} {\hat\omega_{opt}}^T G^T ||^2\\
    &\le 3 \epsilon + || \hat G  \hat\omega_{opt} {\hat\omega_{opt}}^T \hat G^T - \hat G \hat\omega_{opt} {\hat\omega_{opt}}^T G + \hat G \hat\omega_{opt} {\hat\omega_{opt}}^T G -  G  \hat\omega_{opt} {\hat\omega_{opt}}^T G^T ||^2\\
    &\le 3 \epsilon + || (\hat G - G)  \hat\omega_{opt} {\hat\omega_{opt}}^T \hat G^T ||^2 + || (\hat G - G)  \hat\omega_{opt} {\hat\omega_{opt}}^T G^T ||^2\\
    & \le 3 \epsilon + || (\hat G - G)  \hat\omega_{opt} {\hat\omega_{opt}}^T \hat G^T - (\hat G - G)  \hat\omega_{opt} {\hat\omega_{opt}}^T G^T + (\hat G - G)  \hat\omega_{opt} \hat\omega_{opt}^T  G^T ||^2 + d^2|| \hat G - G||^2 ||G ||^2\\
    & \le 4 \epsilon  + ||(\hat G - G)  \hat\omega_{opt} {\hat\omega_{opt}}^T  (\hat G - G) ||^2 + || (\hat G - G)  \hat\omega_{opt} {\hat\omega_{opt}}^T G^T ||^2\\
    & \le 5 \epsilon + d^2 || \hat G - G||^4 \\
    & \le 6 \epsilon 
\end{align*}

This means if we find an $\epsilon-$approximate solution to the system with the estimated values, we obtain a $\kappa_{min}$ within $6\epsilon$ of the optimal. 
\end{proof}

Finally, we present the proof of Lemma \ref{lem:grec}:
\begin{proof}
Recall that when the decision-maker fixes $\omega$, it receives samples of the form $x + G \omega$. We note this can be used to recover the matrix $G$. In particular, we show how $d$ rounds, each with $O(\frac{d \mathrm{tr}(\Sigma)}{\epsilon})$ samples, suffices to recover the matrix to squared Frobenius norm $\epsilon$. 
Recall the procedure we propose simply chooses $\omega = e_1,...e_d$, one-hot coordinate vectors in each round. 
We first bound the error in $\hat G$. coordinate-wise:  $\mathbb E[||\hat G_{i,j} - G_{i,j}||^2] \le \frac{\mathbb E[x_i^2]}{n}$. A union bound across coordinates shows that $O(\frac{d^2 \mathrm{tr}(\Sigma)}{\epsilon})$ samples suffice to recover $G$ within squared Frobenius norm $\epsilon$.
\end{proof}

\end{document}